\documentclass{article}
\usepackage{natbib}
\setcitestyle{numbers,square}

\usepackage[preprint]{neurips_2025}

\usepackage[utf8]{inputenc} 
\usepackage[T1]{fontenc}    
\usepackage{hyperref}       
\usepackage{url}            
\usepackage{booktabs}       
\usepackage{amsfonts}       
\usepackage{nicefrac}       
\usepackage{microtype}      
\usepackage{xcolor}         

\usepackage{graphicx}
\usepackage{amsmath}
\usepackage{amssymb}
\usepackage{enumitem}
\usepackage{colortbl}
\usepackage{array}
\usepackage{multirow}
\usepackage{longtable}
\usepackage{makecell}
\usepackage{wrapfig}
\usepackage{graphicx}
\usepackage[ruled]{algorithm2e}
\usepackage{tabularray}
\usepackage{pifont}
\usepackage{subcaption}
\SetKwInput{KwInput}{Input}
\SetKwInput{KwOutput}{Output}
\usepackage{float}
\usepackage{tikz}
\usepackage{bibunits}
\usepackage{setspace}
\definecolor{LightSkyBlue}{RGB}{135 206 250} 

\newcommand{\yzq}[1]{\textcolor[rgb]{0.0, 0.0, 0}{#1}}
\newcommand{\yu}[1]{\textcolor[rgb]{0.0, 0.0, 0.0}{#1}}
\newcommand{\ryn}{\textcolor[rgb]{0,0,0}}
\newcommand{\rynq}{\textcolor[rgb]{0,0,0}}
\newcommand{\rl}{\textcolor[rgb]{0,0,0}}

\title{Unleashing the Potential of Multimodal LLMs for Zero-Shot Spatio-Temporal Video Grounding}

\author{
Zaiquan Yang, \ \ \ 
Yuhao Liu$^{\dagger}$, \ \ \ 
Gerhard Hancke, \ \ \ 
Rynson W.H. Lau$^{\dagger}$
\vspace{1mm}
\\
Department of Computer Science
\\
\vspace{1mm}
City University of Hong Kong
\\
~\texttt{\{zaiquyang2-c, yuhliu9-c\}@my.cityu.edu.hk} \\
~\texttt{\{gp.hancke, Rynson.Lau\}@cityu.edu.hk}
}

\begin{document}

\newcommand{\grayfont}{\color{black!55}}
\newcommand{\LG}[1]{\cellcolor{LightSkyBlue!20}#1}
\newcommand{\TG}[1]{\textcolor{gray}{#1}}
\maketitle

\makeatletter{}\renewcommand*{\@makefnmark}{}
\footnotetext{$^{\dagger}$ Joint Corresponding authors.  \makeatother}



\begin{abstract}
Spatio-temporal video grounding (STVG) aims at localizing the spatio-temporal tube of a video, as specified by the input text query.
In this paper, we utilize multimodal large language models (MLLMs) to explore a zero-shot solution in STVG.
We reveal two key insights about MLLMs: 
(1) MLLMs tend to dynamically assign special tokens, referred to as \textit{grounding tokens}, for grounding the text query; and
(2) MLLMs often suffer from suboptimal grounding due to the inability to fully integrate the cues in the text query (\textit{e.g.}, attributes, actions) for inference. Based on these insights, we propose a MLLM-based zero-shot framework for STVG, which includes novel decomposed spatio-temporal highlighting (DSTH) and temporal-augmented assembling (TAS) strategies to unleash the reasoning ability of MLLMs.
The DSTH strategy first decouples the original query into attribute and action sub-queries for inquiring the existence of the target both spatially and temporally.
It then uses a novel logit-guided re-attention (LRA) module to learn latent variables as spatial and temporal prompts, by regularizing token predictions for each sub-query.
These prompts highlight attribute and action cues, respectively, directing the model's attention to reliable spatial and temporal related visual regions.
In addition, as the spatial grounding by the attribute sub-query should be temporally consistent,
we introduce the TAS strategy to assemble the predictions using the original video frames and the temporal-augmented frames as inputs to help improve temporal consistency.
We evaluate our method on various MLLMs, and show that it outperforms SOTA methods on three common STVG benchmarks.
 The code will be available at \url{https://github.com/zaiquanyang/LLaVA_Next_STVG}.
\end{abstract}

\section{Introduction}
\label{sec:intro}
\yu{Spatio-Temporal Video Grounding (STVG)} aims to localize a target object \ryn{in a video both spatially and temporally, given an input text query. This task is fundamental to many different} applications (\textit{e.g.}, video surveillance and autonomous driving~\cite{zeng2025futuresightdrive}
).
\ryn{However, it is also very challenging as it requires the model to be able to distinguish the target from distractors over time and identify the precise temporal boundary of the action.}
\yu{\ryn{While existing methods handle the STVG task} mainly} in \ryn{a fully-supervised setting}~\cite{garg2025stpro, yang2022tubedetr, wasim2024videogrounding}, which often \ryn{relies} on costly frame-level annotations, 
\yu{\ryn{several} works~\cite{li2023winner, bao2024e3m, yang2022learning, jin2024weakly, kumar2025contextual} attempt to} introduce \yzq{\ryn{the weakly-supervised} or zero-shot setting} \yu{to alleviate the burden of} dense annotations.
\ryn{For example,} E3M~\cite{bao2024e3m} 
\yzq{integrates CLIP~\cite{radford2021learning} and an expectation maximization strategy to optimize spatio-temporal localization} \yu{in a zero-shot \ryn{manner}.} However, CLIP is known to be weak in localization~\cite{zhou2022extract, ghiasi2022scaling} as it simply aligns the global \ryn{representation} of image-text pairs.

\yu{Considering the strong \ryn{capability} of multi-modal large language models (MLLMs)~\cite{liuvisual, li2024llava, lin2024video, chen2024internvl, wang2024qwen2,liu2025language} in cross-modality alignment, several works explore \ryn{the application of MLLMs in the} visual grounding task.}
\ryn{However, they typically} require explicit fine-tuning~\cite{lai2024lisa, rasheed2024glamm, ma2024groma} of MLLMs with additional grounding datasets and \yu{specific} model modifications, which \ryn{can be difficult} to scale and generalize to novel visual data~\cite{cao2024emerging}. 
\ryn{Although some recent works~\cite{cao2024emerging, zhang2025mllms} \yzq{\ryn{have investigated} the attention maps in language models as done in previous ViT and CNN architectures~\cite{zhou2016learning, touvron2021training, gao2021ts},
}}
\yzq{\ryn{they only focus} \yu{on} the generated tokens while neglecting other token components input in the language model (\textit{e.g.}, system tokens \ryn{and} special tokens).} 
\rynq{\ryn{In particular}, 
\yzq{\ryn{we observe that} the special tokens play an important role in structuring the communication between the user input and the language model, and help guide the generation of coherent responses in dialogues.
}}
\begin{figure*}[t]
    \centering
    \includegraphics[width=0.99\textwidth]{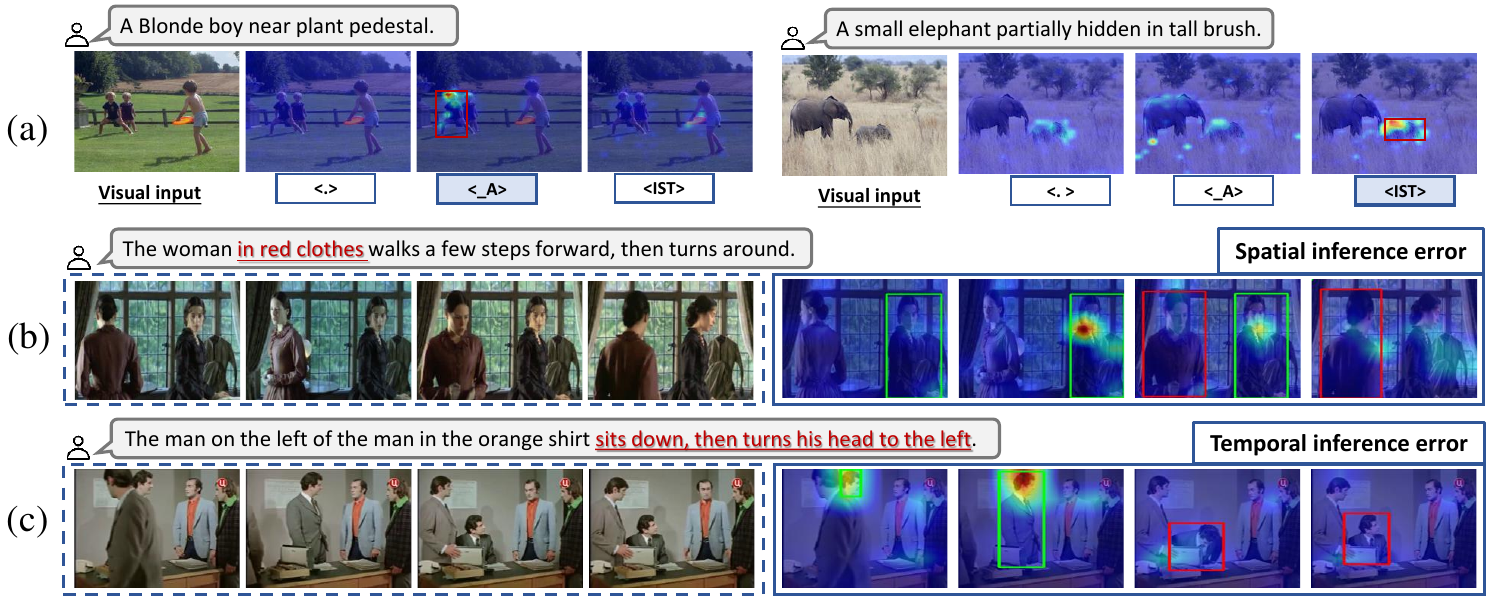}
    \vspace{-1mm}
    \caption{
    (a) The visual attention maps shows that some special tokens (marked as \raisebox{-0.1em}{\includegraphics[width=1.4em, height=0.8em]{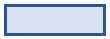}}) can precisely attend to the target region \ryn{of the} input query.
    However, these special tokens, referred to as the \textit{grounding tokens} in our work, underperform on complex STVG, where they often focus on part cues and ignore other cues (marked in red within the input text prompts).  
    Examples (b) and (c) illustrate spatial/temporal grounding errors caused by ignoring the discriminative attribute/action cues. Red and green bounding boxes denote the ground truths and predictions, respectively.
    }
    \label{fig:teaser}
    \vspace{-4mm}
\end{figure*}
%

\yu{With the above observation in mind, we delve deeper and find}
that the special tokens \yzq{following} the input \ryn{instruction have} outstanding grounding ability. 
In particular, a few special tokens are characterized by \ryn{high} \yzq{visual activation}
and \ryn{can attend to the region of interest well}.
\yzq{\ryn{Considering the left sample of Fig.~\ref{fig:teaser}(a) as an example, the special token `\texttt{\_A}'}
provides tangible attention to the \textit{`boy'} referred \ryn{to} by the given query, while in the right sample of Fig.~\ref{fig:teaser}(a), the token `\texttt{IST}' is assigned to \ryn{ground} the \textit{`elephant'}.} These special tokens are referred to as \textit{\textbf{grounding tokens}} in our work.
Despite \ryn{the strong comprehension ability}, the grounding tokens cannot optimally adapt to STVG \ryn{as they tend to ignore} some important cues (\textit{e.g.}, attribute or action) in \ryn{a} complex video query.
%
As shown in Fig.~\ref{fig:teaser}(b), the grounding token fails in \ryn{the spatial grounding by} ignoring the attributes.
Similarly, in Fig.~\ref{fig:teaser}(c), it leads to the failure of temporal grounding \ryn{by} neglecting the action cues of \ryn{the} target.

In this work, 
we first \ryn{conduct a} systematic analysis to probe the special tokens of various MLLMs (Sec.~\ref{subsec:GTI}), \yzq{which demonstrates our \ryn{aforementioned} empirical findings}
and enables zero-shot spatio-temporal video grounding.
To alleviate the problem of neglecting discriminative cues, we propose a novel decomposed spatio-temporal highlighting (DSTH) strategy (Sec.~\ref{subsec:DSH}). Specifically, the language query is decomposed into attribute and action sub-queries, which are utilized as the text input of the MLLM for inquiring the target’s existence \ryn{spatially and temporally}, respectively.
\ryn{We then propose a novel logit-guided re-attention (LRA) module} to highlight the cues in attribute and action sub-queries. For each sub-query, LRA optimizes the learnable latent variable as the (spatial/temporal) prompts via
\yzq{enhancing the positive response generation while suppressing the negative response}.
As a result, the DSTH strategy \ryn{can well adapt} the model to mine faithful visual context and concentrate on \ryn{spatially/temporally} relevant regions. 
In addition, \yzq{to further enhance the temporal consistency during spatial grounding by the attribute sub-query, we develop a temporal-augmented assembling (TAS) strategy (Sec.~\ref{Subsec:TAS}).}
\ryn{The \yu{TAS} strategy} utilizes temporally perturbed frames as input \yu{to} assemble different predictions into final spatial grounding result.
%
%

In summary, our contributions are as follows:
\begin{itemize}[leftmargin=*]
\item We reveal that MLLMs dynamically assign the special tokens for precisely grounding text-related regions.
We identify the special tokens characterized by the salient visual activation for deriving a novel zero-shot STVG framework.
\item We propose a novel test-time tuning strategy named decomposed spatio-temporal highlighting (DSTH). It introduces an innovative logit-guided re-attention (LRA) module to adapt the grounding token for thorough spatio-temporal localization. We also develop a temporal-augmented assembling (TAS) strategy \ryn{to further improve} the robustness of spatial localization.
\item We conduct extensive experiments on various MLLMs to demonstrate our empirical findings and validate the effectiveness of the proposed method. Our method outperforms existing methods by a remarkable margin on three STVG benchmarks.
\end{itemize}

\vspace{-3mm}
\section{Related work}

\noindent \textbf{Spatio-Temporal Video Grounding (STVG)} aims to localize a spatio-temporal tube in a video corresponding to the text query.
Unlike \ryn{image-based visual grounding}~\cite{deng2021transvg, ding2021vision, yang2022lavt, liu2023gres, yang2024boosting}, STVG~\cite{zhang2020does, tang2021human, yang2022tubedetr} presents significant challenges, requiring models to distinguish targets from distractors both spatially and temporally.
Fully-supervised STVG approaches~\cite{tang2021human, wasim2024videogrounding, zhang2020does, su2021stvgbert} have achieved promising results. However, these methods heavily depend on an extensive collection of labor-intensive annotations. 
Several recent works~\cite{li2023winner, jin2024weakly, kumar2025contextual} tackle STVG in \ryn{a} more efficient manner, which only uses coarse video-level descriptions for training.
E3M~\cite{bao2024e3m} leverages pre-trained vision-language \ryn{models and proposes} an expectation maximization framework to optimize spatio-temporal localization.
However, contrastive objective based multimodal models are known to be weak in localization~\cite{zhou2022extract, ghiasi2022scaling} as they simply align the global representations of image-text pairs.

\noindent \textbf{Multimodal Large Language Models (MLLMs)} are capable of handling diverse language and vision tasks. To equip MLLMs with the grounding ability, current works~\cite{munasinghe2023pg, lai2024lisa, li2025llava,yuan2024osprey, zhangomg} construct grounding-oriented supervision data for instruction tuning and \ryn{propose novel} architectural modifications. 
LLaVA-ST~\cite{li2025llava} achieves spatio-temporal understanding by introducing a progressive training strategy consisting of three sequential stages.  
However, \ryn{the large amount of grounding data needed for instruction-tuning imposes high labeling costs. In addition}, \ryn{changing the focus of MLLMs to grounding tasks} can \ryn{degrade the original dialog capabilities \yzq{due to catastrophic forgetting~\cite{zhai2023investigating}}}.
Recent works~\cite{cao2024emerging, jiang2024devils, zhang2025mllms} reveal the inherent perception ability of MLLMs obtained by general instruction-tuning.
%
\ryn{Unlike these works that only focus} on the generated tokens or 
\ryn{in our work, we delve deeper into the \yzq{more} token components and reveal} that the MLLMs always dynamically assign the special tokens following the instruction prompt for attending to the regions of interest.

\noindent \textbf{Test-time Tuning (TTT)} aims to optimize the inference of test samples online.
With the progress of multimodal foundation models~\cite{radford2021learning, chen2023protoclip}), \yzq{TTT} has attracted more attention~\cite{zhou2022learning, jia2022visual, shu2022test, yoonc} as it can learn effective prompts for test samples and well adapt the foundation model for zero-shot applications. 
TPT~\cite{shu2022test} pioneers the study on TTT by minimizing the prediction entropy between each test sample and its augmented views. HisTPT~\cite{zhanghistorical} explores memory learning for test-time prompt tuning by introducing the memory bank. However, the test prompt optimization for \ryn{MLLMs} is barely explored.
\ryn{The most relevant work} to our work is ControlMLLM~\cite{wucontrolmllm}, which optimizes \ryn{the} attention map by taking the referring regions as supervision. 
With the lack of supervision, we propose to learn visual \ryn{prompts} by regularizing \ryn{the} token-level response to instruction input. Our work \ryn{shows} that we can rectify where text \ryn{prompts} attend to by altering the outputs of MLLMs.
%



\vspace{-0mm}
\section{Method}
\label{sec:met}

\subsection{Preliminaries}

\textbf{Task Formulation.}
Given an untrimmed video $V = \{\mathrm{f}_{t}\}_{t=1}^{T_{v}}$ composed of $T_{v}$ image frames and a sentence query $Q$, 
the goal of the STVG \ryn{task} is to localize the spatio-temporal tube of target ${O}_{}= \{{b}_{t}\}_{t=t_s}^{t_e}$ described by $Q$. Here ${b}_{t}$ represents the bounding box of \ryn{the} target in the $t$-th frame, $t_{s}$ and $t_e$
specify the starting and ending \ryn{boundaries} of the target tubelet, respectively. 
The STVG task can be solved \ryn{through two sub-tasks: spatial grounding and temporal grounding}. In our work, we first derive a zero-shot solution for \ryn{the} STVG task by unleashing the strong cross-modal comprehension ability of \ryn{MLLMs}.

\textbf{\ryn{The Setup of MLLMs}}. 
Current MLLMs typically consist of a visual encoder, a projector, and a large language model. 
Specifically, given \ryn{an} image-question pair ${(I, Q)}$ as input,
the image $I$ is \ryn{first} projected into text-aligned visual tokens $\mathrm{T}_{v} = \{\mathrm{v}_1, \dots, \mathrm{v}_{\mathrm{M}}\}$ by visual encoder and projector, and the question $Q$ is converted into text tokens $\mathrm{T}_{q} = \{\mathrm{t}_1, \dots, \mathrm{t}_\mathrm{N_{q}}\}$ by a text tokenizer and embedding layer. Here, $\mathrm{M}$ and $\mathrm{N_{q}}$ denote the numbers of visual tokens and question tokens, respectively.
In practice, MLLMs also introduce system tokens $\mathrm{T_{sys}}$ and some special tokens $\mathrm{T_{role}}$ for instruction-following ability.
In particular, the special tokens play an important role in structuring the well-organized conversation framework. In our work, the special tokens are positioned subsequent to the instruction prompts by the user. They help guide the generation of coherent responses. The number of special tokens $\mathrm{N_{role}}$ often differs among \ryn{different} MLLMs.
As a result, the language model receives concatenated tokens, $\mathrm{T} = \{
\mathrm{t}_1, \dots, \mathrm{t}_\mathrm{N_{sys}};
\mathrm{v}_1, \dots, \mathrm{v}_{\mathrm{M}};
\mathrm{t}_1, \dots, \mathrm{t}_\mathrm{N_{q}};
\mathrm{t}_1, \dots, \mathrm{t}_\mathrm{N_{role}}
\}$, as input. 
\ryn{Appendix~\ref{sec:appendix_tokens_input} provides visual illustration about the} input tokens in MLLMs.

\textbf{Text-to-visual Attention.}
The LLM in MLLMs typically processes the input tokens through \textrm{L} transformer blocks~\cite{liuvisual} with the multi-head attention (MHA) for interactions of different tokens.
Particularly, the text-to-visual attention represents the relationships between the visual and the textual tokens. 
We can derive the text-to-visual attention matrix $A \in \mathbb{R}^{\mathrm{L} \times \mathrm{H} \times \mathrm{N} \times \mathrm{N}}$, where $\mathrm{N} = \mathrm{N_{sys}} + \mathrm{M} + \mathrm{N_{q}} + \mathrm{N_{role}}$. $\mathrm{L}$ and $\mathrm{H}$ denote \ryn{the numbers of layers and heads in the} transformer.
\ryn{Our empirical observation from Fig.~\ref{fig:teaser} is that} the special tokens $\mathrm{T_{role}}$ show  outstanding grounding ability with a global comprehension. 
For the simplicity of the following analysis, we omit the affect of different layers and heads by the mean operation.
\ryn{We can then} obtain \yu{the text-to-visual attention matrix}, 
$A_\mathrm{role} = [\mathrm{N_{sys}} + \mathrm{M} + \mathrm{N_{q}} : \mathrm{N},  \mathrm{N_{sys}} : \mathrm{N_{sys}} + \mathrm{M}]$.
\subsection{Grounding Token Identification}
\label{subsec:GTI}
\begin{wrapfigure}{r}{0.60\textwidth} 
    \vspace{-5mm}
    \centering
    \includegraphics[width=0.60\textwidth]{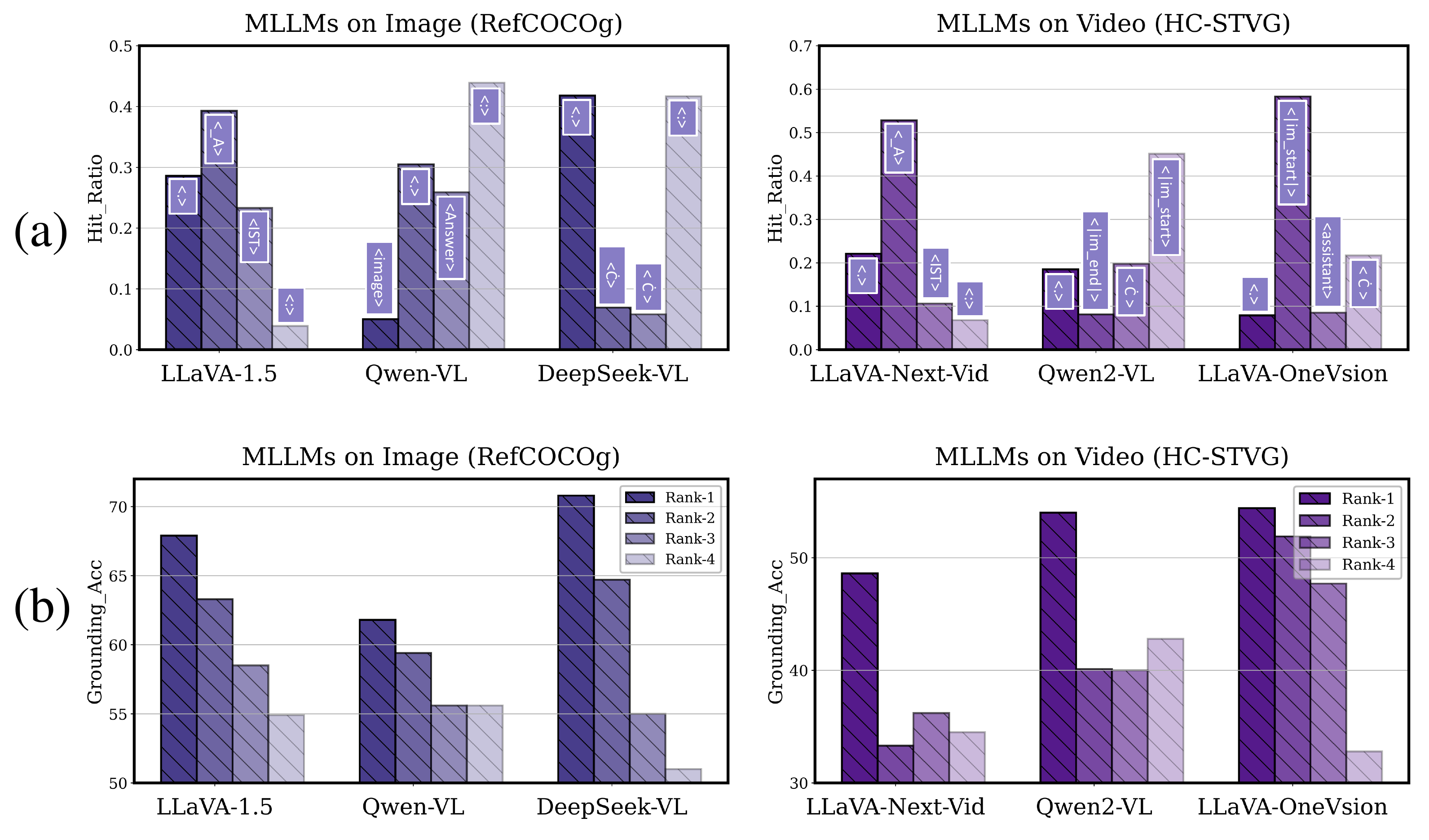}
    \vspace{-5.0 mm}
    \caption{
    In (a), the results show the frequency with which different special tokens (such as \textbf{`\_A'}, \textbf{`.'}) have superior grounding ability than other tokens, \textit{i.e.}, hit\_ratio.
    In (b), the results represent the grounding accuracy of tokens ranked by their visual activation degrees.
    For each MLLM, we select four tokens for visualization.
    }
    \label{fig:grounding_token}
    \vspace{-5mm}
\end{wrapfigure}
In this section, we conduct a pilot study to \ryn{quantitatively analyze the grounding ability of the special tokens.}
Without loss of generality, we randomly \ryn{select} a subset of \texttt{1,000} image-text pairs from \ryn{the} RefCOCOg~\cite{hu2016segmentation} validation set for image MLLMs analysis, and a subset of \texttt{1,000} video-text pairs from \ryn{the} HC-STVGv2~\cite{tang2021human} dataset for video MLLMs analysis.
Particularly, we choose three typical MLLMs (\textit{i.e.}, \texttt{LLaVA-1.5}, \texttt{Qwen-VL}, \texttt{Deepseek-VL}) for \ryn{the study on image} input and three MLLMs (\textit{i.e.},~\texttt{LLaVA-Next-Video}, \texttt{Qwen2-VL}, \texttt{LLaVA-OneVision}) for \ryn{the study on video} input. 
%
\yu{We have two key findings \ryn{from our studies}.}

\textbf{MLLMs dynamically assign the special tokens to attend to the text-related regions.} 
To demonstrate the finding, we first define \ryn{the \textit{\textbf{attention ratio}} of each special token} as the ratio of maximum attention within the ground-truth bounding box $b_{gt}$ to that outside it. For example, the \ryn{attention ratio of the special token \texttt{`\_A'}} can be computed as:
\begin{equation}
\label{att_ratio}
    \mathrm{R_{att}^{\texttt{\_A}}}= \frac{\max(A_\mathrm{role}^{\texttt{\_A}} \odot f_{\mathrm{B2M}}\left({b_{gt}}\right))} 
    {\max 
    \left(
    A_\mathrm{role}^{\texttt{\_A}} \odot \left(1 - f_{\mathrm{B2M}}\left({b_{gt}}\right) \right) 
    \right)}
    , \hspace{0.25in} A_\mathrm{role}^{\texttt{\_A}}\in \mathbb{R}^{1\times \mathrm{M}} \text{,}
\end{equation}
where $f_{\mathrm{B2M}}$ denotes the function \yu{that transforms a bounding box into its corresponding binary mask.}
\yu{Here, a} higher ratio indicates \ryn{a} better target grounding ability. 
\ryn{Given} a test sample, we can identify the token yielding the highest attention ratio as the superior token for grounding.
\ryn{The \textbf{\textit{hit ratio}} of a token is then} defined as the frequency of being the superior token for grounding across all test samples.
\ryn{Fig.~\ref{fig:grounding_token}(a)} shows the hit ratio of four special tokens in each MLLM.
Notably, the fixed token does not consistently exhibit the best grounding ability for different samples. For example, the highest hit ratio achieved by token \texttt{`.'} in  \texttt{LLaVA-1.5} is not more than $50\%$. 
In addition, for different MLLMs, the token at a fixed position does not always yield the best localization performance. For \ryn{example},
\yzq{the last special token \texttt{`:'}, which is
adopted in \ryn{a} previous work~\cite{zhang2025mllms}}, obtains a high hit ratio in \texttt{Qwen-VL}, but its hit ratio is quite low in \texttt{LLaVA-1.5}.
This observation holds for both image and video inputs.

\textbf{The special token with higher visual activation tends to show superior grounding performance.}
Since the superior grounding tokens vary across different samples and MLLMs, 
identifying them for grounding is a problem that needs to be resolved when ground truth is unavailable as a prior.
\ryn{With} further analysis, we reveal that the superior token for grounding tends to show higher visual activation.
%
\yzq{For each sample, we rank the special tokens according to the maximum value of visual attention and then evaluate their grounding accuracy by selecting the proposal with the highest attention value.}
Following the paradigm in previous works~\cite{subramanian2022reclip, han2024zero}, we extract box proposals using a detector and evaluate the grounding accuracy by the Acc$@0.5$ metric.
\ryn{Fig.~\ref{fig:grounding_token}(b) shows the results. 
We can see that the} grounding accuracy decreases as the rank of visual activation \ryn{reduces} (from left to right). 
\yzq{\ryn{This} supports our hypothesis that the special token with higher visual activation tends to show superior grounding ability.}
%
We \ryn{have also observed} that models with better comprehension \ryn{possess better grounding ability overall}. For example, the special tokens in \texttt{LLaVA-OneVision} achieve better grounding performance than those of the other MLLMs. 

\begin{figure*}[t]
    \centering
    \includegraphics[width=0.98\textwidth]{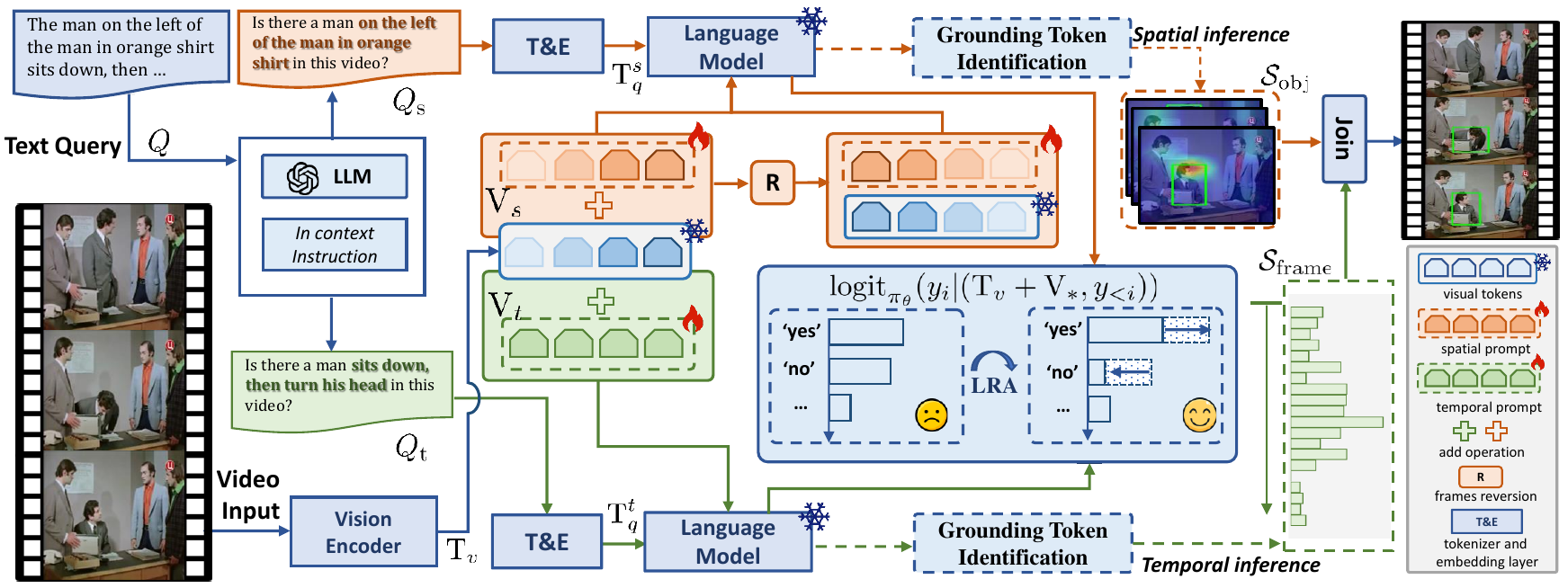}
    \caption{\ryn{Overview} of the proposed approach for zero-shot STVG. 
    Given a video-text pair, we first decompose the text $Q$ into spatially and temporally related sub-queries, $Q_{\mathrm{s}}$ and $Q_{\mathrm{t}}$.
    The text prompt tokens converted from  $Q_{\mathrm{s}}$ and $Q_{\mathrm{t}}$ are then concatenated with visual tokens $\mathrm{T}_{v}$ for spatial and temporal \ryn{inferences}, respectively. 
    In addition, we introduce learnable variables as visual prompts and optimize them by the logit-guided re-attention (LRA) module.
    \yzq{For spatial grounding, we also develop a temporal-augmented assembling (TAS) strategy by reversing the frames to enhance temporal consistency.}
    After optimization, we obtain the \ryn{object track score} $\mathcal{S}_{\mathrm{obj}}$ and \ryn{frame score} $\mathcal{S}_{\mathrm{frame}}$ based on the grounding token identification.
    The final prediction is derived by joining  $\mathcal{S}_{\mathrm{obj}}$ and $\mathcal{S}_{\mathrm{frame}}$.
    }
    \label{fig:MAIN}
    \vspace{-5mm}
\end{figure*}

\textbf{A straightforward solution for zero-shot STVG.} 
Inspired by the grounding ability of special tokens subsequent to the text prompt, we refer to them as \textit{\textbf{grounding tokens}} in our work. By identifying the special tokens characterized by high visual activation, we derive a strong training-free framework for STVG. 
Specifically, given the video-text pair $(V, Q)$, we first extract the object track proposals $\mathcal{O}_{pro} = \{{O}_{1}, \dots, {O}_{P} \}$ from the video $V$, as done in previous works~\cite{bao2024e3m, kumar2025contextual}, \yu{where $P$ denotes the number of proposals, and $O_p = \{{b}_{t}^{\prime} \}_{t=1}^{t=T_{v}}$ is the \ryn{set of bounding boxes of} the $p$-th proposal in $T_{v}$ frames.}
\ryn{Based on the foregoing findings, we then} select the token with the highest attention value for locating the target object, \ryn{and denote the text-to-image attention matrix of the selected token} as $A_\mathrm{g} \in \mathbb{R}^{1 \times M}$. 
As a result, we obtain each object track score by computing the maximum attention value inside each object track as:
\begin{equation}
\label{alignment_score}
    \mathcal{S}_{\mathrm{obj}}= \{
    s^{o}_{1} \text{, } s^{o}_{2} \text{, } \cdots \text{, } s^{o}_{P}
    \}, \text{ with } 
    {s}^{o}_{p}=\max(A_\mathrm{g} \odot f_{\mathrm{B2M}}\left({O_{p}}\right)) \text{,}
\end{equation}
where $\odot$ denotes element-wise multiplication. We choose the track with \ryn{the} highest score as spatial prediction $O_{\textrm{pred}}^{\prime}$. In a similar manner, we can compute the \yzq{frame score
}
$\mathcal{S}_{\mathrm{frame}} = \{
    s^{t}_{1} \text{, } s^{t}_{2} \text{, } \cdots \text{, } s^{t}_{\mathrm{T}_{v}}
    \}$. 
By selecting the top-$\mathrm{K}$ frames with the highest scores from the temporal prediction $\mathcal{S}_{\mathrm{frame}}$, we obtain the final spatio-temporal prediction $O_{\textrm{pred}}^{\prime}= \{{b}_{t}^{\prime} \}_{t=t_{s^{\prime}}}^{t_{e^{\prime}}}$, where $t_{s^{\prime}}$ and  $t_{e^{\prime}}$ denote the starting and ending boundary \ryn{predictions}.
Though simple, \ryn{this} solution has achieved comparable \ryn{or even superior performance} than current zero-shot methods.
For example, based on \texttt{LLaVA-OneVision} model, this solution achieves $23.3$\% on the m\_vIoU metric on HC-STVGv1 dataset, which outperforms previous SOTA result (19.1\%) by E3M~\cite{bao2024e3m}.

\subsection{Decomposed Spatio-Temporal Highlighting}
\label{subsec:DSH}
Despite \ryn{a} strong performance by the above solution, \ryn{these} grounding tokens often neglect some important cues during spatio-temporal localization especially when processing complex video \ryn{queries}. 
As shown in Fig.~\ref{fig:teaser}(b), the attribute cue \textit{`in red clothes'} is overlooked during inference, which causes the incorrect spatial localization. We observe that the language query often contains attribute and action descriptions \ryn{of} the target \yu{object}, which are beneficial for spatial and temporal localization, respectively.
To this end, we propose a novel decomposed spatio-temporal highlighting (DSTH) strategy in our framework, \ryn{\yzq{which aims at highlighting the attributes/ actions cues in language query and enhances the spatial/temporal reasoning, respectively.}} 
%

%
\noindent \textbf{Generation of target-related cues.}
For \yu{comprehensive} spatial and temporal reasoning, it is essential to extract the \ryn{attribute and action descriptions from the original query $Q$ as textual cues of the target.
Here,} we leverage the strong in-context capability of the LLM~\cite{achiam2023gpt} to extract \ryn{attribute} and action descriptions. 
Following previous works~\cite{yang2024boosting, han2024zero}, 
we construct a prompt with general instructions and in-context task examples.
As shown in Fig.~\ref{fig:MAIN}, we feed the prompt into LLM, and then these related descriptions $Q_{\mathrm{s}}$ and $Q_{\mathrm{t}}$ are generated through the LLM completion. 
More implementation details can be found in Appendix~\ref{sec:appendix_Cues_Generation}.

%

\noindent \textbf{Spatio-temporal prompt learning.}
With the decomposed \ryn{attribute and action} descriptions as cues for spatial and temporal reasoning, the \yu{next question} is how to efficiently direct the model to focus on the corresponding visual regions.
Reliable responses in visual question answering (VQA) necessitate careful attention to the relevant visual context as \ryn{pointed out by} previous studies~\cite{xiao2024can,prasad2023rephrase}.
Inspired by this, we innovatively propose regularizing the response to the questions constructed from the \ryn{attribute and action} descriptions for adjusting the visual attention of MLLMs.
Specifically, we first transform the descriptions into interrogative queries by a fixed template to inquire the existence of \ryn{the} target.
As shown in Fig.~\ref{fig:MAIN}, \yzq{from the original text input $Q$, we can obtain the \ryn{attribute} description `\textit{a man \ryn{on} the left of the man in the orange shirt}'}, which is further transformed into interrogative sub-query $Q_{\mathrm{s}}$: {`\textit{Is there a man \ryn{on} the left of the man in the orange shirt in this video?}' 
} 
for spatial inquiry. 
In a similar way, we can obtain the interrogative sub-query $Q_{\mathrm{t}}$ for temporal inquiry.
These interrogative sub-queries will be taken as \ryn{input instructions} of MLLMs for response generation.

\noindent \yu{\textbf{Logit-guided re-attention.}} \yu{With extracted sub-queries above,} \ryn{we then propose a novel logit-guided re-attention (LRA) module} to regularize the token prediction during response generation.
\ryn{Taking the spatial sub-query as an example, we first} initialize a learnable variable $\mathrm{V}_{s}$ with the same shape as \ryn{the} visual tokens $\mathrm{T}_{v}$, and then add it to $\mathrm{T}_{v}$ as the visual input of \ryn{the} language model. Sub-query $Q_{\mathrm{s}}$ is \yzq{converted as text prompt tokens $\mathrm{T}_{q}^{s}$} by the text tokenizer and embedding layer. Given the input \ryn{token} sequence, the next token probability prediction over the vocabulary set $\mathcal{V}$ is formulated as:
\begin{equation}
    \mathrm{p}_{y} = \mathrm{exp} \left( \mathrm{logit}_{\pi_\theta} (y_{i} | (\mathrm{T}_{v} + \mathrm{V}_{s}, \mathrm{T}_{q}^{s}, y_{<i})) \right) ,
\label{eq:llm}
\end{equation}
where ${\pi_\theta}$ denotes the parameter of \ryn{the} language model and is frozen in our work.
$\mathrm{logit}_{\pi_\theta}$ is the log probability of the generated token at time step $i$.
$y_{<t}$ denotes the text tokens sequence prior to prediction time step $i$. 
We define the optimization objective by contrasting the \ryn{probabilities} of positive token \textit{`yes'} and negative token \textit{`no'}:
\begin{equation}
    \mathcal{L}_{s} = 1 - \mathrm{exp} \left( 
    \mathrm{logit}_{\pi_\theta} (y_{i}^{\textit{yes}} | (\mathrm{T}_{v} + \mathrm{V}_{s}, \mathrm{T}_{q}^{s}, y_{<i}))  - \mathrm{logit}_{\pi_\theta} (y_{i}^{\textit{no}} | (\mathrm{T}_{v} + \mathrm{V}_{s}, \mathrm{T}_{q}^{s}, y_{<i}))
    \right).
\label{eq:loss}
\end{equation}
During the inference process, we conduct backpropagation to optimize the learnable variable $\mathrm{V}_{s}$ as \ryn{the} spatial prompt. The process is iterated $\mathrm{N}_{ep}$ times by test-time tuning paradigm.
By enhancing the positive response towards the sub-query $Q_{\mathrm{s}}$, \yu{we can prompt the MLLM to effectively mine target-related contextual information during VQA, which in turn highlights} the attribute cues in the original text.
Similarly, we can obtain the temporal prompt $\mathrm{V}_{t}$ by optimizing the temporal inference.

\noindent \textbf{Joint inference.}
Based on the spatial and temporal visual \ryn{prompts}, we derive the attention \ryn{maps} $A_\mathrm{g}^{S}$ and $A_\mathrm{g}^{T}\in \mathbb{R}^{\mathrm{T}_{v} \times \mathrm{h} \times \mathrm{w}}$ of \ryn{the} special token with high visual activation for spatial and temporal \ryn{predictions}, where $h$ and $w$ \ryn{denote} \ryn{the token numbers of the height and width}. 
\ryn{From Eq.~\ref{alignment_score}, we obtain} the object track score $\mathcal{S}_{\textrm{obj}}$ and the temporal score $\mathcal{S}_{\mathrm{frame}}$ based on $A_\mathrm{g}^{S}$  
 and $A_\mathrm{}^{T}$, respectively.
Finally, we integrate the predictions $O_{\textrm{pred}}^{\prime}= \{b_{t}^{\prime}\}_{t=t_{s^{\prime}}}^{t_{e^{\prime}}}$ as the spatio-temporal grounding result.

\vspace{-1pt}
\subsection{Temporal-augmented Assembling}
\label{Subsec:TAS}
The \ryn{attribute} sub-query, which provides \ryn{a} static state description, exhibits temporal independence for spatial grounding. In other words, the spatial grounding by the \ryn{attribute} sub-query should be temporally consistent for temporal augmentation (\textit{e.g.}, reversing the order of video frames).
\yzq{However}, there exists \yzq{temporal inconsistency} when introducing temporal augmentation in current MLLMs.
Here, we \ryn{propose} a metric to measure the temporal consistency.
\ryn{By denoting the spatial attention maps} before and after reversing the order of input frames as $A_\mathrm{g}^{S}$ and $\tilde{A}_\mathrm{g}^{S}$, \ryn{the}
temporal consistency can be measured as:
\begin{equation}
    {S}_{\mathrm{cons}} = \mathrm{max} 
    \{s_{1} \text{, } \cdots \text{, } s_{P}\}
    , \hspace{0.1in}
    s_{p} = \mathrm{max} \left( 
    ( 
    A_\mathrm{g}^{S} \odot f_{\mathrm{B2M}} \left( {O_{p}} \right) 
    ) 
    \odot 
    ( 
    \tilde{A}_\mathrm{g}^{S} \odot f_{\mathrm{B2M}} \left( {O_{p}} \right)
    )
    \right)
    \text{,}
\label{eq:TS}
\end{equation}
\begin{wrapfigure}{r}{0.48\textwidth} 
    \centering
    \vspace{-4mm}
    \includegraphics[width=0.48\textwidth]{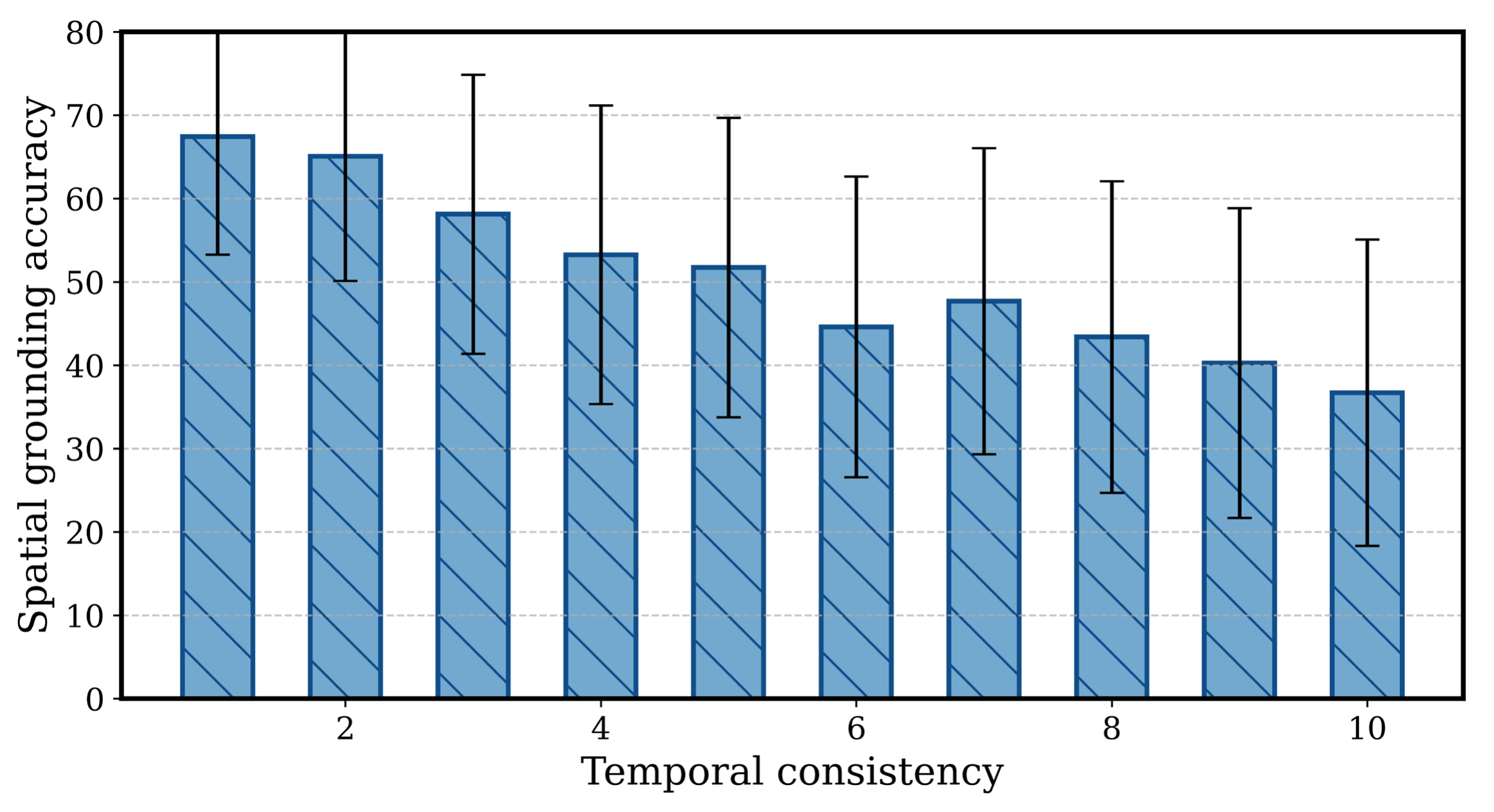} 
    \vspace{-5mm} %
    \caption{
    Spatial grounding accuracy of different groups of samples on \ryn{the} HC-STVGv1 dataset. These groups are ranked by descending temporal consistency.
    }
    \label{fig:TS}
    \vspace{-4mm}
\end{wrapfigure}
\ryn{\yu{where a} higher} ${S}_{\mathrm{cons}}$ indicates better temporal consistency.
\yu{We then }divide the testing samples into ten groups in \ryn{descending} order of temporal consistency.
\ryn{Fig.~\ref{fig:TS} shows the average grounding accuracy of each group samples.}
The results demonstrate a pronounced association between temporal consistency and spatial grounding performance.
The temporal inconsistency tends to cause worse spatial localization. 

To this end, we integrate a temporal-augmented assembling (TAS) strategy in our framework. 
As shown in Fig.~\ref{fig:MAIN}, we perform a frame-level reversion operation on the visual tokens and the spatial prompt simultaneously, and then optimize the spatial prompts for the input of the original frames and the temporal-augmented input, respectively.
During inference, we derive the spatial prediction by assembling the attention maps of temporal-augmented \yzq{input frames}. The proposed TAS strategy alleviates the effect of temporal inconsistency and improves the robustness of spatial grounding.

\section{Experiments}
\label{sec:exp}


\begin{table*}[t] 
\caption{
Quantitative comparison on HCSTVG (v1\&v2) and VidSTG (Declarative) benchmarks.  
}
\vspace{0mm}
\centering
\small    
\renewcommand\arraystretch{1.15}
\setlength{\tabcolsep}{2pt}
\resizebox{0.99\textwidth}{!}
{
    \begin{tabular}{c | l | ccc | ccc | ccc}
      \toprule [1.2pt]
      \multirow{2}{*}{\textbf{Sup}}
      & \multirow{2}*[0.0ex]{\textbf{{Method}}}
      & \multicolumn{3}{c|}{\textbf{HCSTVG-v1}} 
      & \multicolumn{3}{c|}{\textbf{HCSTVG-v2}} 
      & \multicolumn{3}{c}{\textbf{VidSTG (Declarative)}} 
      \\
      &
      &m\_vIoU &vIoU@0.3 &vIoU@0.5
      &m\_vIoU &vIoU@0.3 &vIoU@0.5
      &m\_vIoU &vIoU@0.3 &vIoU@0.5
      \\
      \midrule [0.8pt]
      
      \multirow{5}{*}{\textbf{Full}}
      
      & TubeDETR~\cite{yang2022tubedetr} \textsubscript{\textcolor{gray}{[CVPR2022]}} 
      &32.4 &49.8 &23.5
      &36.4 &58.8 &30.6
      &30.4 &42.5 &28.2
      \\
      & STCAT~\cite{jin2022embracing} \textsubscript{\textcolor{gray}{[NeurIPS2022]}}  
      &35.0 &57.7 &30.0
      &-- &-- &-- 
      &33.1 &46.2 &32.6 
      \\
      & CSDVL~\cite{lin2023collaborative} \textsubscript{\textcolor{gray}{[CVPR2023]}} 
      &36.9 &62.2 &34.8
      &38.7 &65.5 &33.8
      &33.7 &47.2 &32.8
      \\
      & CG-STVG~\cite{gu2024context} \textsubscript{\textcolor{gray}{[CVPR2024]}} 
      &38.4 &61.5 &36.3
      &39.5 &64.5 &36.3
      &34.0 &47.7 &33.1
      \\
      
      \midrule [0.5pt]
      
      \multirow{5}{*}{\textbf{Weak}} 
      
      & WINNER~\cite{li2023winner} \textsubscript{\textcolor{gray}{[CVPR2023]}} 
      &14.2 &17.2 &6.1
      &-- &-- &-- 
      &11.6 &14.1 &7.4
      \\
      & VEM~\cite{jin2024weakly} \textsubscript{\textcolor{gray}{[ECCV2024]}} 
      &14.6 &18.6 &5.8
      &-- &-- &--
      &14.5 &18.6 &8.8
      \\
      & CoSPaL~\cite{kumar2025contextual} \textsubscript{\textcolor{gray}{[ICLR2025]}} 
      &22.1 &31.8 &19.6
      &22.2  &31.4 &18.9 
      &16.0 &20.1 &13.1
      \\
      & STPro~\cite{garg2025stpro} \textsubscript{\textcolor{gray}{[CVPR2025]}} 
      &17.6 &27.0 &12.9
      &20.0 &31.1 &14.6
      &15.5 &19.4 &12.7
      \\
      \midrule [0.5pt]
      
      \multirow{7}{*}{\textbf{ZS}}

      & RedCircle~\cite{shtedritski2023does} \textsubscript{\textcolor{gray}{[CVPR2023]}} 
      &9.2 & 7.8 & 1.6
      &-- &-- &--
      &8.6 &7.6 &0.9
      \\
      & ReCLIP$^{}$~\cite{subramanian2022reclip}  \textsubscript{\textcolor{gray}{[ACL2022]}} 
      &14.4 &18.3 &4.9 
      &-- &-- &--
      &14.2 &17.5 &7.9
      \\
      & E3M~\cite{bao2024e3m} \textsubscript{\textcolor{gray}{[ECCV2024]}} 
      &19.1 &29.4 &10.6
      &-- &-- &--
      &16.2 &20.5 &11.9
      \\
      & \LG{Ours$_{ \text{~LLaVA-Next-Video-7B}}$}
      &\LG{20.4} &\LG{33.6} &\LG{12.4}
      &\LG{23.6} &\LG{36.8} &\LG{15.5}
      &\LG{16.6} &\LG{26.8} &\LG{11.1}
      \\
      & \LG{Ours$_{ \text{~ShareGPT4Video-8B}}$}
      &\LG{20.0} &\LG{32.2} &\LG{10.9}
      &\LG{24.4} &\LG{38.9} &\LG{15.4}
      &\LG{17.1} &\LG{27.8} &\LG{11.6}
      \\
      & \LG{Ours$_{ \text{~Qwen2-VL-7B}}$}
      &\LG{23.6} &\LG{39.0} &\LG{14.4}
      &\LG{25.6} &\LG{40.5} &\LG{17.1}
      &\LG{17.0} &\LG{27.4} &\LG{11.4}
      \\

      & \LG{Ours$_{ \text{~LLaVA-OneVision-7B}}$}
      &\LG{24.8} &\LG{41.5} &\LG{16.3}
      &\LG{27.7} &\LG{44.7} &\LG{19.5}
      &\LG{18.0} &\LG{29.8} &\LG{12.2}
      \\
      \bottomrule[1.2pt]
    \end{tabular}
}
\vspace{-3mm}
\label{tab:HCTSVG_SOTA}
\end{table*}

\subsection{Settings}
\label{sec:exp_setting}
\noindent \textbf{Datasets.} 
We evaluate on three video benchmark datasets: HCSTVG-v1, HCSTVG-v2~\cite{tang2021human}, and VidSTG~\cite{zhang2020does}. We provide detailed introduction about them in Appendx~\ref{sec:appendix_Datasets}.
%

\noindent \textbf{Implementation Details.}
We adopt G-DINO~\cite{liu2024grounding} and SAM2~\cite{ravi2024sam} for detection and tracking tubelet generation, 
and use GPT-4o to decompose the original query sentence into spatial and temporal sub-queries.
We consider four widely-used video MLLMs:  
\texttt{LlaVA-Next-Video-7B}~\cite{li2024llava}, 
\texttt{Qwen2-VL-7B}~\cite{wang2024qwen2}, 
\texttt{ShareGPT4Video-8B}~\cite{chensharegpt4video}, 
\texttt{LlaVa-OneVision-7B}~\cite{li2024llava} 
for demonstrating the efficiency of our method.
\rl{Refer to} Appendix~\ref{sec:appendix_Implementation} for more details.

\noindent \textbf{Evaluation \rl{Metrics}.}
We follow the standard evaluation protocol~\cite{yang2022tubedetr, kumar2025contextual} and use $\mathrm{m\_vIoU}$, and $\mathrm{vIoU@R}$ to assess the performance of spatio-temporal grounding. 
Specifically, let $\mathrm{S}_{i}$, $\mathrm{S}_{u}$ denote the intersection and union between the predicted and ground-truth frames. The $\mathrm{vIoU}$ is computed by $\frac{1}{\mathrm{S}_{u}} \sum_{t\in \mathrm{S}_{i}} \textbf{IoU}(b_{t}^{\prime}, b_t)$, where $b_{t}^{\prime}$ and $b_t$ denote the detected and ground-truth bounding box at frame $t$, respectively.
The $\mathrm{m\_vIoU}$ represents the $\mathrm{vIoU}$ averaged over all testing samples, \rl{and} $\mathrm{vIoU@R}$ denotes the proportion of data samples in the testing subset with $\mathrm{vIoU}$ greater than the threshold $\mathrm{R} \in \{0.3, 0.5\}$.

\vspace{-1mm}
\subsection{Performance Comparison}
\label{subsec:sota}

\noindent \textbf{Quantitative Comparison.}
Tab.~\ref{tab:HCTSVG_SOTA} presents a comparison of our method against 11 methods from three categories, including zero-shot, weakly-supervised, and fully-supervised methods. 
Specifically, we compare our approach with existing zero-shot SOTA approaches 
(\textit{e.g.}, E3M~\cite{bao2024e3m}, ReCLIP~\cite{subramanian2022reclip}, RedCircle~\cite{shtedritski2023does}).
Our method consistently outperforms these methods by a remarkable margin on all benchmarks.
Based  on \texttt{LLaVA-Next-Video-7B}, our method outperforms  E3M by \rl{$\mathrm{4.2}$\% on $\text{vIoU}@0.3$ and $\mathrm{1.8}$\% on $\text{vIoU}@0.5$}.  
When integrated into \rl{the} better \texttt{LLaVA-OneVision-7B} model, the \rl{corresponding} improvements reach $\mathrm{12.1}$\% and $\mathrm{5.7}$\%.
\begin{wrapfigure}{r}{0.65\textwidth} 
    \centering
    \vspace{-2mm}
    \includegraphics[width=0.65\textwidth]{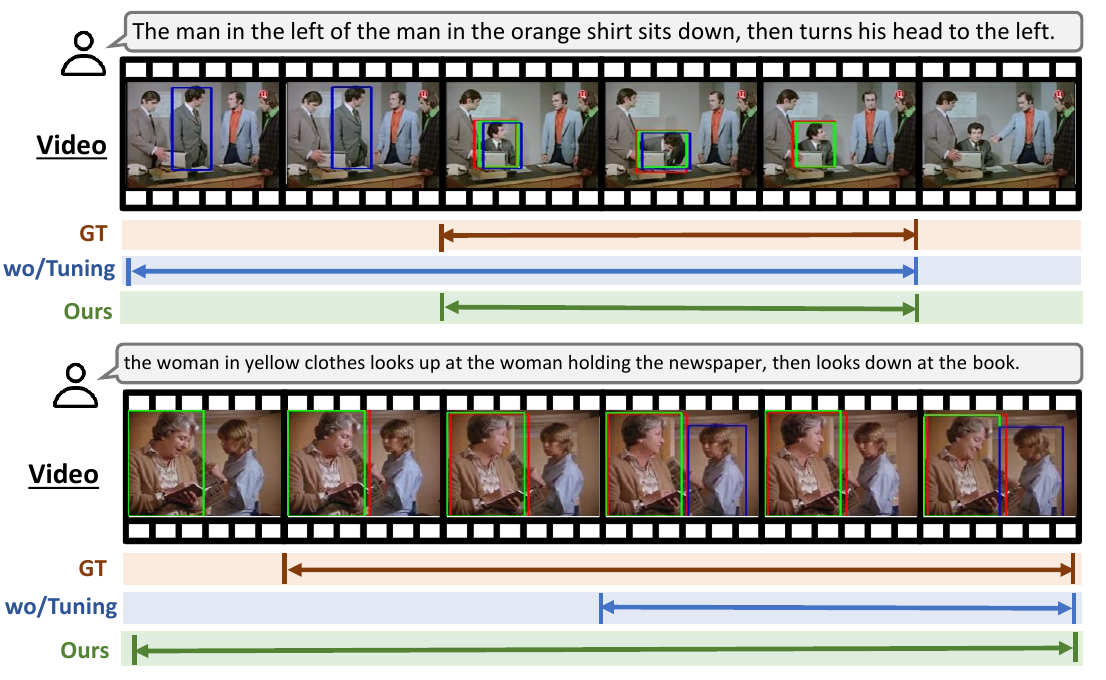} %
    \vspace{-4mm} %
    \caption{
    Qualitative results on \rl{the} HC-STVGv1 test set. Better spatio-temporal grounding results (green) are \rl{obtained when the DSTH strategy is being used} for optimization.
    }
    \label{fig:Qualitative_Vis}
    \vspace{-3mm}
\end{wrapfigure}
\rl{This} shows that our framework can \rl{adapt to various  MLLMs well, and better performances can be achieved by using} better MLLMs.
\yzq{Even on the VidSTG dataset, which contains fewer action cues related to the target and thus makes temporal grounding particularly challenging, our framework still outperforms the previous SOTA \rl{methods} overall. This demonstrates the strong generalization capability of our method.}
%
Besides, 
our method even surpasses current SOTA weakly-supervised methods on most metrics. 
For example, on the HCTSVG-v2 benchmark, our method outperforms CoSPaL~\cite{kumar2025contextual} by a margin \rl{of} $\mathrm{5.5}$\% on the $\text{m\_vIoU}$ metric. 
Furthermore, compared to the fully-supervised methods, our method \rl{can still} achieve comparable results, which further validates the superiority of our approach. We provide quantitative comparison on the VidSTG (Interrogative) and image benchmarks in Appendix~\ref{subsec:appendix_VidSTG_Interrogative}

\noindent \textbf{Qualitative Comparison.}
We present qualitative results in Fig.~\ref{fig:Qualitative_Vis}. 
In the example below, before introducing the DSTH optimization strategy, the model neglects the attribute cues in the language query and suffers from the spatial grounding error. 
By highlighting the attribute cues of \rl{the} target, our method can direct the MLLMs toward reliable visual context and improve spatial localization.
%
%

\subsection{Ablation Study}
\label{subsec:ablation}

In this section, based on HC-STVGv1 dataset, we analyse the effect of different proposed components when integrated into $\texttt{LlaVa-Next-Video}$ and $\texttt{LlaVa-OneVision}$.
We also conduct extensive ablation experiments \rl{with the} hyper-parameters based on $\texttt{LlaVa-Next-Video}$.

\noindent \textbf{Component analysis.}
In Tab.~\ref{tab:Component}, we first average the attention maps of all special tokens (\textbf{\textit{1st}} row) as the baseline.
This solution has achieved outstanding performance.
Next, in \rl{the} \textbf{\textit{2nd}} row, we integrate the selection of the superior token introduced in grounding token identification (GTI). It brings consistent improvements on different MLLMs. The results show that identifying the superior special token can effectively unleash the powerful comprehension ability of MLLMs.
\rl{We them} validate the efficiency of the decomposed spatio-temporal highlighting strategy (\textbf{\textit{3rd}} \rl{and \textbf{\textit{4th}} rows}). 
$\textbf{S}_{\text{tune}}$ and $\textbf{T}_{\text{tune}}$ denote adopting the prompt learning in spatial and temporal inferences, respectively.
It is demonstrated that the MLLMs can be directed to \rl{focus on the spatial/temporal related regions better} when highlighting the \ryn{attribute/action} cues with proper prompts. 
When utilizing prompt learning on the two sub-tasks, the final performance can be further refined (\textbf{\textit{5th}} row). 
\rl{In addition, we also find that} the DSTH strategy can improve the grounding performance especially for the less efficient MLLMs (\textit{e.g.}, \texttt{LLaVA-Next-Video}). 
Even for the MLLMs (\textit{e.g.}, \texttt{LLaVA-OneVision}) \rl{with} strong temporal comprehension ability, \rl{performance improvement can still be} achieved by the test-time optimization.
%
%
\begin{wraptable}{h}{0.7\textwidth}
\vspace{-3mm}
    \caption{
        \rl{Component} ablation on \texttt{LLaVA-Next-Video} and \texttt{LLaVA-OneVision}.
    }
    \small  
    \renewcommand\arraystretch{1.1}
    \centering
    \vspace{-2pt}
    \setlength\tabcolsep{2pt}%
    \resizebox{\linewidth}{!}
    {
    \begin{tabular}{p{0.65cm} p{0.5cm} p{0.4cm} p{0.6cm}|ccc|ccc} 
        \toprule
        \multirow{2}{*}{\textbf{GTI}}
        & \multirow{2}{*}{\text{$\textbf{S}_{\text{p}}$}}
        & \multirow{2}{*}{\text{$\textbf{T}_{\text{p}}$}}
        & \multirow{2}{*}{$\text{TAS}$}
        & \multicolumn{3}{c|}{\texttt{LLaVA-Next-Video}}  & \multicolumn{3}{c}{\texttt{LLaVA-OneVision}}
        \\
         &  &  & & m\_vIoU  & vIoU@0.3 & vIoU@0.5  & m\_vIoU  & vIoU@0.3 & vIoU@0.5
         \\
        \midrule
        \ding{55} & \ding{55} & \ding{55}  & \ding{55}
        &15.2  & 25.1 & 8.5
        &21.3  &36.1  & 12.6      \\
         \ding{51} & \ding{55} & \ding{55}  &  \ding{55}
        &16.3 & 26.6 & 9.3 
        &23.3 & 38.8 & 15.1     \\
         \ding{51}  &  \ding{51}  & \ding{55} & \ding{55}
        & 18.0  & 30.1 & 10.1
        &24.1   & 40.3   & 16.0 \\
        \ding{51}  &  \ding{55}  &\ding{51}  &   \ding{55}
        & 18.4 & 29.5 & 10.1  
        & 23.8 & 39.7 &  15.5     \\
        \ding{51}  & \ding{51} & \ding{51} &   \ding{55} 
        & 19.9 & 32.1 & 11.9 
        &24.3 & 40.7 &  16.0  \\
        \ding{51}  & \ding{51} & \ding{51}  & \ding{51} 
        & 20.4 & 33.6 & 12.4 
        &24.8 & 41.5 &  16.3     \\
        \bottomrule
        \end{tabular}
    }
    \label{tab:Component}
    \vspace{-5mm}
\end{wraptable}
%
Finally, we introduce the temporal-augmented assembling (TAS) strategy (\textbf{\textit{6th}} row). The overall performance can be further improved by refining the spatial localization.

%
\begin{figure*}[t]
    %
    %
    \begin{minipage}[t]{1.00\textwidth}
        \centering
        \vspace{-5mm}
        \includegraphics[width=1.00\linewidth]{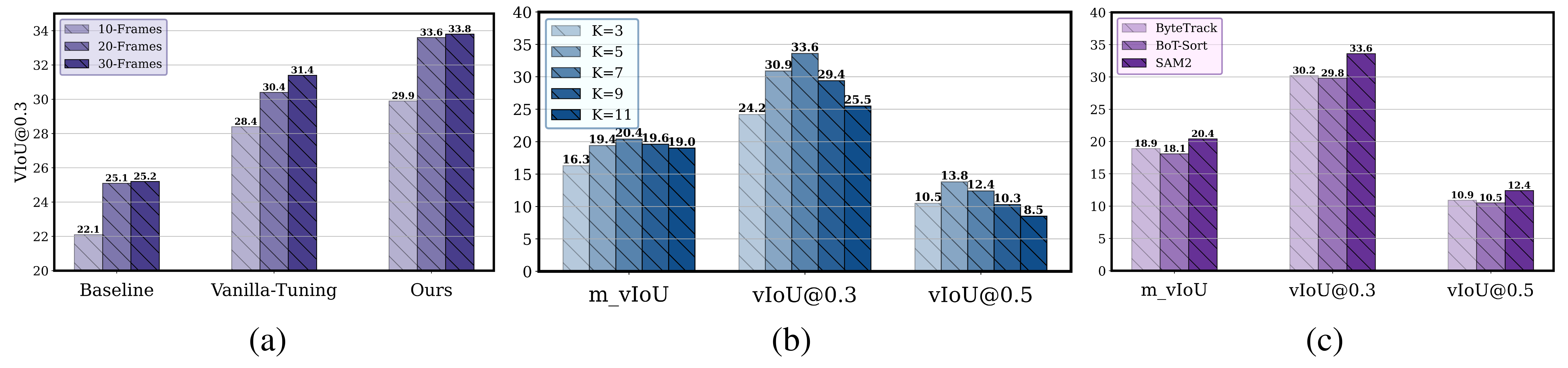}
        \vspace{-6mm}
        \caption{(a) Comparison \rl{on the number of frames $\mathrm{N}_{f}$ as input, using vIoU@$0.3$}. 
        (b) Ablation on the number of selected frames $\mathrm{K}$ during temporal prediction.
        (c) Ablation on different trackers.}
        \label{fig:appendix_Frames_Trackers}
    \end{minipage}%
    \vspace{2mm}
    \hfill
    %
    %
    %
\vspace{-6mm}
\end{figure*}
%
%

\noindent \textbf{Effect of input frames $\mathrm{N}_{f}$.}
\rl{Fig.~\ref{fig:appendix_Frames_Trackers}(a) analyzes} the effect of \rl{using} different numbers of video frames $\text{N}_{f}$ as input.
We can see that more \rl{input frames} provide richer visual context for model inference and lead to \rl{a better overall performance}. 
Besides, the performance becomes slightly peaked as the number of input frames increases. To balance \rl{between} performance and efficiency during inference, we uniformly sample 20 frames as the visual input of MLLMs. 
\yzq{In Fig.~\ref{fig:appendix_Frames_Trackers}, we also evaluate\texttt{Vanilla-Tuning}, a solution that directly applies test-time optimization to the entire text query $Q$ instead of decomposing it into \rl{attribute/action} sub-queries.
We observe that the performance of \texttt{Vanilla-Tuning} is inferior to our method.
It demonstrates that decomposing the text query into \rl{attribute/action} sub-queries can help facilitate the intensive spatio-temporal comprehension of MLLMs.}

\noindent \textbf{Ablation \rl{on the} number $\mathrm{K}$ of predicted frames.}
\rl{Fig.~\ref{fig:appendix_Frames_Trackers}(b) analyzes} the effect of predicted frame numbers $\mathrm{K}$ for temporal grounding based on the HCSTVG-v1 dataset. 
Here, we consider selecting the optimal frame number from the set $\{3, 5, 7, 9, 11\}$. In general, when $\mathrm{K}$ is set as $7$, the optimal result can be obtained. \rynq{
\yzq{Thus, we select top-$7$ frames as \rl{the} prediction during temporal grounding.}
}

\noindent \textbf{Trackers ablation.}
\rl{Fig.~\ref{fig:appendix_Frames_Trackers}(c) analyzes} the effect of different trackers. Besides SAM2~\cite{ravi2024sam}, a foundation model for tracking, 
we also consider two other tracking models (\textit{i.e.}, ByteTrack~\cite{zhang2022bytetrack} and BoTSort~\cite{aharon2022bot}) for analysis.
It is reasonable that when a stronger tracking model (\textit{e.g.}, SAM2) is \rl{used} for generating spatio-temporal tubelets, \rl{we can obtain better STVG performances}.
Besides, when suboptimal tracking models are used, our method can still achieve comparable or better \rl{performances} than the current SOTA methods, \yzq{which shows the generalization of our approach.}


\section{Conclusion}
\label{sec:conclusion}
In this paper, we \rl{have presented} a novel MLLM-based zero-shot framework for the spatio-temporal video grounding (STVG) task.
Our approach is initiated by identifying the grounding capability of special tokens in widely used MLLMs during response generation.
%
To leverage and unleash the comprehension ability of MLLMs, we \rl{have proposed} the decomposed spatio-temporal highlighting (DSTH) strategy. 
It first decomposes the text query into attributes and actions sub-queries.
\rl{It then employs a logit-guided re-attention (LRA) module} to sharpen the spatial/temporal visual context comprehension.
We \rl{have also proposed the} temporal-augmented assembling (TAS) strategy to alleviate the effect of temporal inconsistency.
%
Extensive experiments conducted on three STVG benchmarks demonstrate the effectiveness of our proposed framework.
%

\yzq{
Our approach does have limitations. 
For example, \yu{it may struggle to}  process long videos well due to high computational consumption caused by MLLMs. 
\yu{As a future work, we would like to consider incorporating token pruning and key \rl{frame} selection techniques into the model design.}}

\clearpage





\newpage
\newpage

\medskip
{
\bibliographystyle{splncs04}
\bibliography{reference}
}
\end{document}